\pdfoutput=1

\documentclass[11pt]{article}

\usepackage{EMNLP2023}

\usepackage{times}
\usepackage{latexsym}

\usepackage[T1]{fontenc}

\usepackage[utf8]{inputenc}

\usepackage{microtype}

\usepackage{inconsolata}

\usepackage[utf8]{inputenc} 
\usepackage[T1]{fontenc}    
\usepackage{hyperref}       
\usepackage{url}            
\usepackage{booktabs}       
\usepackage{amsfonts}       
\usepackage{nicefrac}       
\usepackage{microtype}      
\usepackage{xcolor}         

\usepackage{hyperref}
\usepackage{url}
\usepackage{graphicx}
\usepackage{amsmath}

\usepackage{caption}
\usepackage{subcaption}

\usepackage{amssymb}
\usepackage{booktabs}

\usepackage{nameref}

\usepackage{pgfplots}
\usepackage{pgfplotstable}

\usepgfplotslibrary{groupplots}
\usepgfplotslibrary{statistics}
\usetikzlibrary{patterns}
\pgfplotsset{every tick label/.append style={font=\footnotesize},compat=1.17}
\pgfplotsset{compat = newest}

\usepackage{ulem}
\definecolor{CREL1}{HTML}{135ea0}
\definecolor{CREL2}{HTML}{b61651}
\definecolor{CREL3}{HTML}{b38600}

\definecolor{C1}{HTML}{1E88E5}
\definecolor{C2}{HTML}{D81B60}
\definecolor{C3}{HTML}{FFC107}
\definecolor{C4}{HTML}{004D40}
\definecolor{C5}{HTML}{D55E00}
\definecolor{C6}{HTML}{785EF0}
\definecolor{C7}{HTML}{16E7CF}

\newcommand{\mytilde}{\raise.17ex\hbox{$\scriptstyle\mathtt{\sim}$}}
\newcommand{\dogball}{$\langle \textrm{dog, chase, ball} \rangle$}
\newcommand{\balldog}{$\langle \textrm{ball, chase, dog} \rangle$}

\newcommand{\astrohorse}{$\langle \textrm{astronaut, ride, horse} \rangle$}
\newcommand{\horseastro}{$\langle \textrm{horse, ride, astronaut} \rangle$}
\newcommand{\svo}{$\langle \textrm{S, V, O} \rangle$} 

\usepackage{varioref}
\usepackage{hyperref}       
\usepackage{cleveref}
\crefname{section}{\S}{\S\S}
\Crefname{section}{\S}{\S\S}
\crefname{table}{Table}{Tables}
\crefname{figure}{Figure}{Figures}
\crefname{algorithm}{Algorithm}{}
\crefname{equation}{eq.}{}
\crefname{appendix}{Appendix}{}
\crefformat{section}{\S#2#1#3}

\title{Training Priors Predict\\ Text-To-Image Model Performance}

\author{Charles Lovering \quad Ellie Pavlick\\
Department of Computer Science \\
Brown University\\
{\texttt{\{first\}}}\_{\texttt{\{last\}}}@brown.edu 
}

\begin{document}

\pgfplotsset{
only if/.style args={entry of #1 is #2}{
/pgfplots/boxplot/data filter/.code={
\edef\tempa{\thisrow{#1}}
\edef\tempb{#2}
\ifx\tempa\tempb
\else
\def\pgfmathresult{}
\fi
}
}
}

\pgfplotsset{
only if both/.style args={entry of #1 is #2 and #3 is #4}{
/pgfplots/boxplot/data filter/.code={
\edef\tempa{\thisrow{#1}}
\edef\tempb{#2}
\edef\tempc{\thisrow{#3}}
\edef\tempd{#4}
\ifx\tempa\tempb\else\def\pgfmathresult{}\fi
\ifx\tempc\tempd\else\def\pgfmathresult{}\fi
}
}
}

\maketitle

\begin{abstract}
Text-to-image models can often generate some relations, i.e., ``astronaut riding horse'', but fail to generate other relations composed of the same basic parts, i.e., ``horse riding astronaut''. These failures are often taken as evidence that models rely on training priors rather than constructing novel images compositionally. This paper tests this intuition on the stablediffusion 2.1 text-to-image model. By looking at the subject-verb-object (SVO) triads that underlie these prompts (e.g., ``astronaut’’, ``ride’’, ``horse’’), we find that the more often an SVO triad appears in the training data, the better the model can generate an image aligned with that triad. Here, by aligned we mean that each of the terms appears in the generated image in the proper relation to each other. Surprisingly, this increased frequency also diminishes how well the model can generate an image aligned with the flipped triad. For example, if ``astronaut riding horse'' appears frequently in the training data, the image for ``horse riding astronaut’’ will tend to be poorly aligned. Our results thus show that current models are biased to generate images with relations seen in training, and provide  new data to the ongoing debate on whether these text-to-image models employ abstract compositional structure in a traditional sense, or rather, interpolate between relations explicitly seen in the training data.
\end{abstract}

\section{Introduction}

Whether neural networks extrapolate beyond their training data is an open question. Part of the debate hinges on the mechanism that supports models' generalization: is it an abstract combinatorial structure of the type that characterized good-old-fashioned-AI \cite{fodor1998}, or rather a ``mix-and-match'' strategy born out of patterns seen in training \cite{lake2018generalization,chaabouni-etal-2020-compositionality}?

Whichever mechanism models are using, both text \citep{openai2023gpt4,ouyang2022training}  and image \citep{ramesh2022hierarchical,rombach2021highresolution,saharia2022photorealistic,yu2022scaling} generation models push far beyond the bounds of what was assumed possible only a few years ago. Anecdotal examples of models writing working code, authoring poetry, and generating images \cite{bubeck2023sparks} make it hard to deny that they possess some mechanism for compositionality. However, failures of such models consistently crop up \cite{gary2022astronaut,mckenzie2022inverse,mckenzie2022round1,mckenzie2022round2}. Moreover, the increasing scale of these models’ training datasets and costs makes it difficult to run controlled experiments, and thus most debates about how models achieve their apparent compositional behavior rely on speculation or in-principle arguments. 

In this work, we analyze the relationship between items seen in training and a model's ability to generalize, in order to better understand what types of mechanisms might underlie the observed behavior.
We focus on the stablediffusion 2.1 text-to-image model. Such models can often generate many relations, i.e., ``astronaut riding horse'', but fail to generate atypical relations composed of the same basic parts, i.e., ``horse riding astronaut''. These failures are often taken as evidence that the models rely on training priors rather than constructing novel images using more systematic mechanisms. By and large, our results support this intuition. Looking at the subject-verb-object relations that underlie these prompts, we find that the increased frequency of a subject-verb-object relation in the training data improves how well the model can generate an image aligned with that relation, but  this increased frequency diminishes how well the model can generate an image aligned with the flipped relation (object-verb-subject). For example, the more common a relation like ``astronaut riding horse'' is in the training data, the worse ``horse riding astronaut’’ is generated. These results demonstrate that the relations seen during training have a significant impact on what the model is able to successfully generate. We also find that frequencies of individual terms, like how often the subject has been seen in any context, have a significant impact on alignment. Together, these results strongly suggest the model uses an underlying ``mix-and-match'' mechanism to support generalization, rather than using a more abstract combinatorial process of the type traditionally assumed to be necessary for models with the observed capabilities.

\section{Experimental Design}
In this section, we formulate our primary questions as hypotheses about how the estimated counts of subject-verb-objects in the training data impact text-to-image prompt and image alignments. 

\begin{figure}[ht!]
    \centering
    \includegraphics[width=0.3\linewidth]{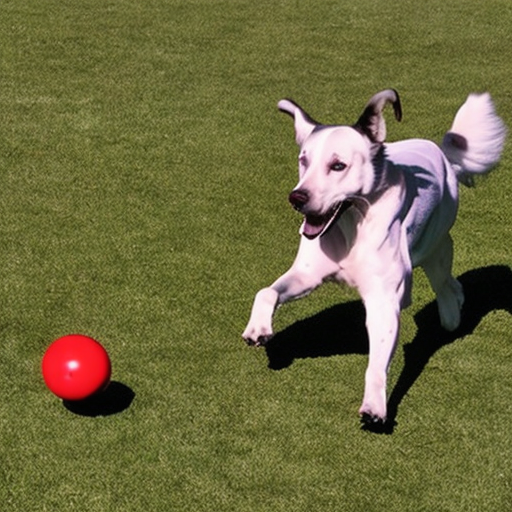}
    \includegraphics[width=0.3\linewidth]{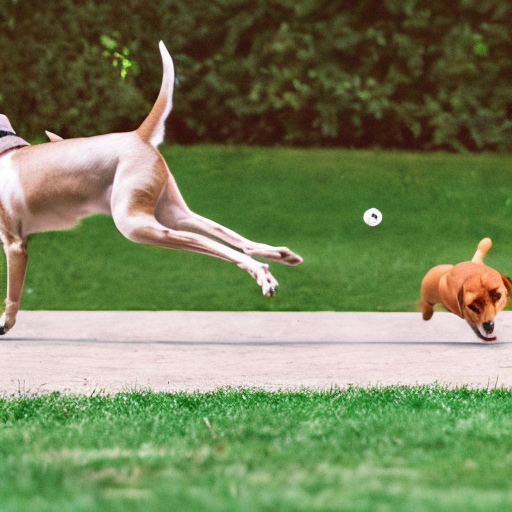}
\caption{\textbf{Example generations.} Left: ``dog chasing a ball''; Right: ``ball chasing a dog.'' These examples were generated by stablediffusion 2.1 (and are cherry-picked out of 4 generations.)
}\label{fig:generation}
\end{figure}

\subsection{The Problem}
Our aim is to characterize when text-to-image models successfully generate relations like in \cref{fig:generation}:\\

\begin{tabular}{ll}
A dog chasing a ball. & $\langle \textrm{dog, chase, ball} \rangle$ \\
A ball chasing a dog. & $\langle \textrm{ball, chase, dog} \rangle$ \\
\end{tabular}
\\

Each prompt has an underlying triad: a subject $s$, a verb $v$, and an object $o$.\footnote{Note that in the paper, our relations are such that the subject is always the semantic agent and the object is the semantic patient. Thus, throughout, we use the terms ``subject'' and ``agent'' interchangeably, and likewise use ``object'' and ``patient'' interchangeably.} For example, for the sentence ``A dog chasing a ball,'' $s = $``dog,'' $v = $``chase,'' and $o = $``ball''. For shorthand, we denote this triad  as $\langle s, v, o \rangle$. We construct our dataset to include both $\langle s, v, o \rangle$ and $\langle o, v, s \rangle$-- we call the more frequently appearing triad the default triad, and the other, the flipped triad.

\paragraph{Notation} To help track various counts, we introduce some notation. This notation is all with respect to a given prompt formatted from a triad $\langle s=S, v=V, o=O \rangle$.\\

\noindent\begin{tabular}{lp{5cm}}
SVO  & Est. count of the relation  \\
OVS & Est. count of the flipped relation \\
{\textit{X}xx}, {x\textit{X}x} $\dots$ & Est. count of \textit{X} in the given role \\
\end{tabular}
\\

\noindent Here, {\textit{X}xx}, {x\textit{X}x} $\dots$ refer to the counts of individual entities that appear in that slot in any context \textit{except} for the counts covered by SVO or OVS.\footnote{That is, the triads that contribute to the count Sxx are disjoint from the triads that contribute to the count of SVO. This is done so that the terms are not algebraically tied, violating the independence assumptions needed in our regressions.} For example, {Sxx} refers to the count of triads with the subject \textit{S} and any verb and object. Analogously, {xxS} refers to the count of triads with the term \textit{S} as the object; {Sxx}, {Oxx}, {xVx}, {xxS}, {xxO} are all similarly defined. These terms are all estimated (parsed) counts from the training data of stablediffusion 2.1.\footnote{In \cref{app:sec:assumptions} we do a deeper dive checking that the (noisy) parsed counts we use reflect the actual counts.} 

For each frequency, we use a log-transformed value in our regressions and plots. For example, for SVO, we use $\log_{10}(\textrm{SVO} + 1)$ s.t. the frequencies range from $[0, \infty]$.

\subsection{Hypotheses}\label{sec:hyp}
We test two primary hypotheses, below.

\begin{quote}
    \noindent\texttt{Forward:} Increased SVO causes better alignment for $\langle S, V, O \rangle$. \\
    \noindent\texttt{Backward:} Increased OVS causes worse alignment for $\langle S, V, O \rangle$.
\end{quote}




\subsection{Statistical Methods}
Though we present results across different views of our dataset, we focus on a single multiple regression to test our hypotheses.
\begin{center}
        Alignment $\sim$ SVO + OVS + \\Sxx + xVx + xxO + Sxx + xxO\footnote{This \textit{R} syntax is meant to show that we see how the alignment of the prompts can be fit by the counts SVO and OVS while being controlled by the individual term counts.}
\end{center}
\noindent Specifically, we are interested in which of the above counts have a significant effect on alignment, controlling for the other counts. We detail how we measure alignment below in \cref{sec:dataset}. In short, we crowdsource $N=5$ ratings per image-prompt pair. There are a number of different options with different tradeoffs on how to handle the multiple ratings per image. If we consider the raters as random effects, we control for raters' variation, but we are no longer able to report $p-$values. Happily, this decision did not meaningfully impact the results, so the regressions use disaggregated ratings (no averaging) without random effects. See alternate formulations in  \cref{app:tab:exp2:controls} and  \cref{app:tab:exp2:average} in the Appendix. For visual acuity, our plots use average alignments.

Our data and data collection break independence assumptions. The individual terms (the S, V, and O) are all individually frequent. Furthermore, we bootstrapped our search to find more relations with high-frequency counts. Both of these decisions mean the relations in our dataset are not randomly sampled from among the population of possible relations. Furthermore, as part of the curation, we manually edit and select prompts that are ``drawable''--i.e., one could in principle draw it, even if the relation is not natural. For example, we filter out relations like ``I love it'' and ``we help child.''
Lastly, the default and flipped versions of a prompt are not independent. The symmetry across the diagonals of \cref{fig:dataset} highlights this point. Splitting the dataset into two partitions, as we do in \cref{sec:exp3}, removes this problem as we run the regressions separately.\footnote{\cref{fig:dataset} plots highlight that the triad frequencies (SVO and OVS) are correlated with the individual frequencies (Sxx, Oxx, \dots). Moreover, the SVO and OVS frequencies between a default and a flipped triad are inverted. To control these dependencies, when we analyze the full dataset in \cref{sec:exp3}, we split default and flipped pairs into different subsets.}

\section{Dataset}\label{sec:dataset}

We use the text-to-image model stablediffusion 2.1 because its training data, code, and model weights are public. We parse its training set, LAION \citep{schuhmann2022laion}\footnote{We use the subset with an aesthetic score 4.5+ as this is what stablediffusion 2.1 was trained on.}, and create a curated dataset of about 769 triads, counts, prompts, generated images, and alignment ratings. 
Later sections (\cref{sec:exp1,sec:exp2,sec:exp3}) use different views of this dataset. The dataset is released in the linked url.\footnote{Dataset: \url{https://docs.google.com/spreadsheets/d/1hcMEMqSX_0regA-s0UplvzrNX2Ho0jHNeIIq-iVoQPI/edit?usp=sharing}}

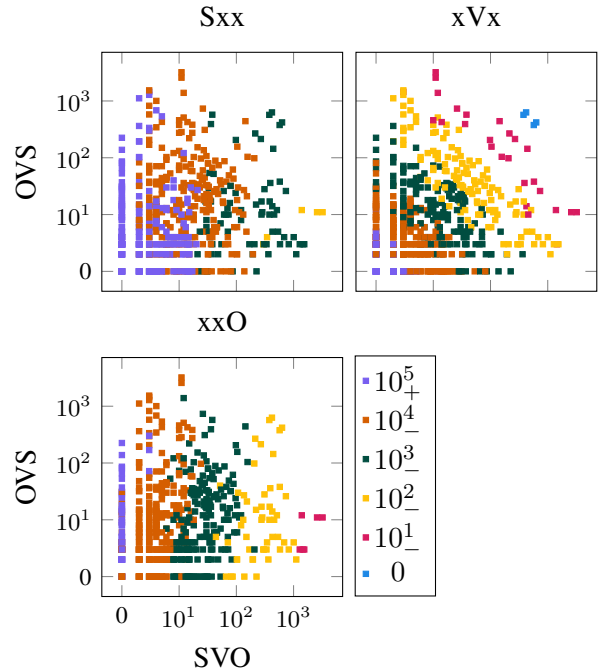
\begin{figure}[ht!]
    \centering 
\begin{tikzpicture}
\begin{groupplot}[group 
style={columns=3},
height=4.75cm,
width=4.75cm,
xtick={0,1,2,3,4},
xticklabels={0,$10^1$,$10^2$,$10^3$},
ytick={0,1,2,3,4},
yticklabels={0,$10^1$,$10^2$,$10^3$},
xlabel=SVO,
ylabel=OVS,
group style={
    group size=2 by 2,
    horizontal sep=5pt,
    vertical sep=25pt,
    x descriptions at=edge bottom,
    y descriptions at=edge left},
]
\nextgroupplot [
title=Sxx
]
\addplot [scatter, 
table/y=OVS,
table/x=SVO,
table/meta=Sxx,
point meta=explicit,
mark=square*, mark size=1pt,
only marks,
scatter/classes={
0={C1},
1={C2},
2={C3},
3={C4},
4={C5},
5={C6}
}
] table {tikz/dataset.dat};
\nextgroupplot [
title=xVx
]
\addplot [scatter, 
table/y=OVS,
table/x=SVO,
table/meta=xVx,
point meta=explicit,
mark=square*, mark size=1pt,
only marks,
scatter/classes={
0={C1},
1={C2},
2={C3},
3={C4},
4={C5},
5={C6}
},
] table {tikz/dataset.dat}
;

\nextgroupplot [
title=xxO,
legend style={at={(1.05, 0.0)},anchor=south west},
legend columns=1,
reverse legend
]
\addplot [scatter, 
table/y=OVS,
table/x=SVO,
table/meta=xxO,
point meta=explicit,
mark=square*, mark size=1pt,
only marks,
scatter/classes={
0={C1},
1={C2},
2={C3},
3={C4},
4={C5},
5={C6}
},
] table {tikz/dataset.dat};

\legend{$0$, $10^1_-$,$10^2_-$,$10^3_-$,$10^4_-$,$10^5_+$};

\end{groupplot}
\end{tikzpicture}
    \caption{\textbf{Dataset Samples Plotted by SVO (x-axis) and OVS (y-axis)}.
    These plots each show the (same) dataset colored by different frequency counts: Sxx, xVx, xxO. The legend details the coloring. \textcolor{C6}{$10^5_+$ examples} are seen that many times or more. \textcolor{C5}{$10^4_-$ examples} are seen less than {$10^4$} but more than $10^3$ times. The other colors are analogously defined. 
    \textbf{As SVO increases along the x-axis, the individual frequencies Sxx and xxO decrease;} (the colors match the gradient of the legend.) It seems that words that appear most frequently in specific relations appear less frequently in others. xVx decreases as both SVO and OVS increase. Thus, \textbf{the individual terms and the triad frequencies are (negatively) correlated}. Each plot is symmetric across the diagonal (not by color) because for each relation we have the default and flipped prompt.\footnotemark } \label{fig:dataset}
\end{figure}

\paragraph{Desiderata} Testing our hypotheses requires collecting relations from the training data, creating a prompt for each relation, generating an image for each prompt, and then measuring how well each prompt and image aligns. Because our dataset is only for evaluation a large dataset is unnecessary. Moreover, because we want to have multiple ratings for each prompt-image pair, a smaller dataset is best. The primary difficulty in constructing such a dataset is finding relations with varying count statistics. These constraints guide some of the decisions we made when collecting our dataset.

\paragraph{Measuring Alignment} We measure the performance of the text-to-image models by asking people how well the prompts and the generated images align. We use SurgeAI \footnote{https://www.surgehq.ai/} to run these experiments; \cref{figure:interface} in \cref{app:sec:label} is a screenshot of the interface. The alignment scores were collected across a 5-point Likert scale from ``Strongly Disagree'' to ``Strongly Agree.'' In our plots and results, we standardize the five ratings from 0 to 1. For the sake of analysis, we consider an average alignment score above 0.75 as a successful generation. (This threshold is chosen because 0.75 and 1 correspond to affirmative responses ``Agree'' and ``Strongly Agree,'' as defined by the rating task.)

\paragraph{Dataset Collection and Measuring Frequency}
\footnotetext{Plots for Oxx and xxS are not shown, but, because of the structure of the dataset, they would be colored symmetrically across the diagonal from the default counterpart (Sxx and Oxx) (xxO and xxS).}
We parse LAION \citep{schuhmann2022laion}\footnote{We sample from the subset with an aesthetic score 4.5+ because this is what stablediffusion 2.1 was trained upon.}, a dataset of paired text and image using Spacy for subject-verb-object relations. The parsing is noisy: Our search finds subject-verb-object relations in $10\%$ of (about 1.3 billion) instances.\footnote{This is admittedly a small fraction. We performed significant manual analysis in order to verify that our 10\% sample is not unduly biased, and thus that our conclusions are reliable despite the small sample. See \cref{app:sec:assumptions} for a detailed discussion.} Many instances are fragments, e.g., ``homes with wrap around porches\dots'', with no subject-verb-object relation. After we collect the counts of subject-verb-objects, we sample relations from different frequencies. Next, we manually create prompts for relations that are ``drawable''. \cref{app:sec:collection} further details the initial collection process and sanity checks a number of assumptions that we outline in \cref{sec:limitations}. \cref{fig:dataset} shows the distribution of examples across the different frequencies defined above; Appendix \cref{app:tab:datastats} reports alignment score statistics.

\paragraph{Formatting Prompts} We format the prompts such that the subject is always the agent (the entity doing the action) and the object is always the patient (the entity on the receiving end of the action). Specifically, we format all relations into prompts using the structure: \textrm{ A photograph of a \{subject\} \{verb\} a \{object\}.} ``A photograph of '' is preprended to each example because in pilot studies it narrowed the type of generation to be photographic, avoiding a lot of trivial errors. After adding the prefix, we edit each example to ensure it is grammatical. For instance, for mass nouns there is no ``a'' included: $\langle \textrm{girl, bring, water} \rangle$ yields ``A photograph of a girl bringing water.'' Though we could generate many reasonable prompt variations for each relation, we have a single prompt per relation (and we generate a single image per prompt). Future work could easily expand our dataset along both of these axes.

\section{Experiments}\label{sec:exps}
 We find broad support for both hypotheses (\cref{sec:hyp}), but the strength of the results depends on the relative size of the priors (SVO vs OVS). By partitioning the dataset,\footnote{In the appendix, \cref{app:fig:partitions} visually shows how the dataset is split across the various experiments across \cref{sec:exp1}, \cref{sec:exp2}, \cref{sec:exp3}. \cref{app:tab:datastats} shows the average alignment scores.} we get a more complete picture of how the frequencies impact the generations.\footnote{Appendix \cref{app:tab:exp0} reports the results for the full dataset.}

 \paragraph{Summary of Takeaways of \cref{sec:exp1,sec:exp2,sec:exp3}}
 For relations that have been observed during training, we find that higher SVO counts result in better alignment. For relations that are entirely unseen, we find that OVS has little effect on alignment. This is surprising, as we initially expected that higher OVS (i.e., a stronger prior for the competing relation) would make it harder for the model to generate the unseen relation--e.g., it should be harder to generate ``horse riding astronaut'' if you have seen ``astronaut riding horse'' often. When SVO is greater than OVS, the effects of both SVO and OVS strongly support the \texttt{forward} and \texttt{backward} hypotheses. However, for flipped examples, cases like \balldog\  with the more infrequent ordering, the effects are weaker. That is, when a model is trying to generate a relation that goes against its training prior, we see less strong effects across the board. The increased difficulty (and lower overall alignment scores, Appendix \cref{app:tab:datastats}) of such examples may explain the differences across partitions.
 Lastly, individual terms Oxx and xxS also impact the results: That is, we find that words appear to have typical roles--they tend to be used either as agents or patients \citep{mahowald2022experimentally}, and models show some resistance to generating them in their less-typical roles.


\subsection{\texttt{Forward:} Increased SVO\\ Increases Alignment}\label{sec:exp1}
First, we want to establish whether or not the increasing frequency of relations in the training data increases the alignment scores, e.g., the \texttt{forward} hypothesis. The relations in this experiment are filtered so that the inverse relation is always unseen, e.g., OVS is 0, isolating the effect of SVO. For example, if $\langle \textrm{dog, chase, ball} \rangle$ is in this subset, then $\langle \textrm{ball, chase, dog} \rangle$ is unseen.

\subsubsection{Results}
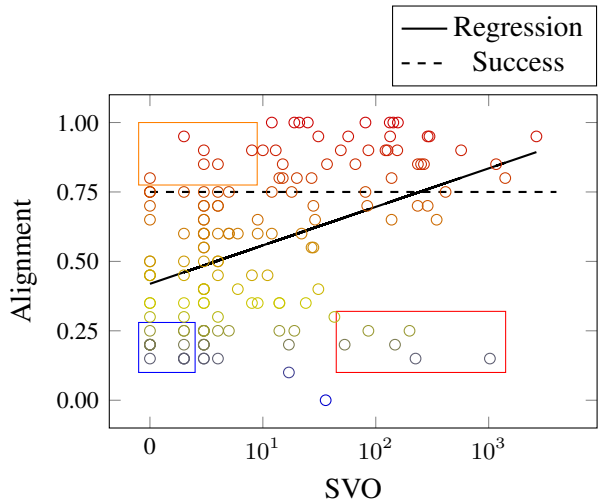
\begin{figure}[ht!]
         \centering 
\begin{tikzpicture}
\begin{axis}[
xtick={0,1,2,3,4},
xticklabels={0,$10^1$,$10^2$,$10^3$},
height=6cm,
width=8cm,
ylabel=Alignment,
xlabel=SVO,
legend style={at={(1.0,1.025)},anchor=south east},
ytick={0.00,0.25,0.50,0.75,1.00},
yticklabels={0.00,0.25,0.50,0.75,1.00},
]

\addplot [no markers, thick, black, table/x=FrequencyLog] table [
y={create col/linear regression={x=FrequencyLog,y=Alignment}}, 
] {tikz/exp1scatter.dat};
\addlegendentry{Regression};
\addplot[mark=none, thick, dashed, black] coordinates {(0,0.75) (3.6 , 0.75)};
\addlegendentry{Success};
\draw (axis cs:-0.1,0.775) rectangle (0.95, 1) [orange];
\draw (axis cs:-0.1,0.1) rectangle (0.4, 0.28) [blue];
\draw (axis cs:1.65,0.1) rectangle (3.15, 0.32) [red];
\addplot [scatter, table/y=Alignment,
table/x=FrequencyLog,
mark=o,only marks, blue] table {tikz/exp1scatter.dat}
;
\end{axis}
\end{tikzpicture}

\caption{\textbf{Alignment improves with Frequency.}  Each alignment score is the mean across 5 ratings. \textcolor{orange}{Top left} samples in \cref{fig:exp1-samples-success-lf}; \textcolor{blue}{Bot left} samples in \cref{fig:exp1-samples-lf}; \textcolor{red}{Bot right} samples in \cref{fig:exp1-samples-hf}.}
         \label{fig:exp1-chart}
     \end{figure}

The results, see \cref{tab:exp12} and \cref{fig:exp1-chart}, support the \texttt{forward} hypothesis. More frequent prompts are better aligned with their outputs. SVO has a $+0.31$ effect over a log scale meaning that for every order of magnitude more, say, \dogball\  has been seen in the training data (holding all else equal) there is an increase of $0.31$ in alignment score. This is substantial, as recall that alignment ranges from 0 to 1.

\begin{table}[ht!]
\centering
\begin{tabular}{lrr}
\toprule
& \multicolumn{2}{c}{SVO Isolated \cref{sec:exp1}}  \\
Term & Effect & \textit{p}-value \\
\midrule
SVO     &     {0.31} &   0.00\\
Sxx      &    0.04   & 0.27\\
xVx      &    {0.12}  &  0.01 \\
xxO      &    0.07  &  0.24\\ 
Oxx      &   {-0.18}  &  0.00 \\
xxS      &   {-0.23}   & 0.00 \\
\bottomrule
\end{tabular}
\caption{\textbf{Regression Coefficients} for Predicting Alignment (when OVS count is 0).
}
\label{tab:exp12}
\end{table}

After $\textrm{SVO} > 10^2$ examples, a  majority of generations are successful (specifically, 18 of 26 prompts were successfully generated with average scores above 0.75.) Several individual terms also impact the results. The effect of Oxx is $-0.18$, meaning the more frequently that the object of relation takes the role of the subject in the training data, the worse the alignment. The same is analogously true for subjects (the effect of xxS is $-0.23$). These results don't discount the \texttt{forward} hypothesis but do suggest that a number of overlapping training priors play a part in the generations, not just the frequency of OVS. The correlation between the frequency and the alignment is 0.421 ($p < 1e-7$). 

\subsubsection{Qualitative Examples}
Firstly, there are some counterexamples to the \texttt{forward} hypothesis: the \textcolor{orange}{orange rectangle} in \cref{fig:exp1-chart} highlights success cases at low-frequencies.

\begin{figure}[ht!]
    \centering
    \includegraphics[width=0.95\linewidth]{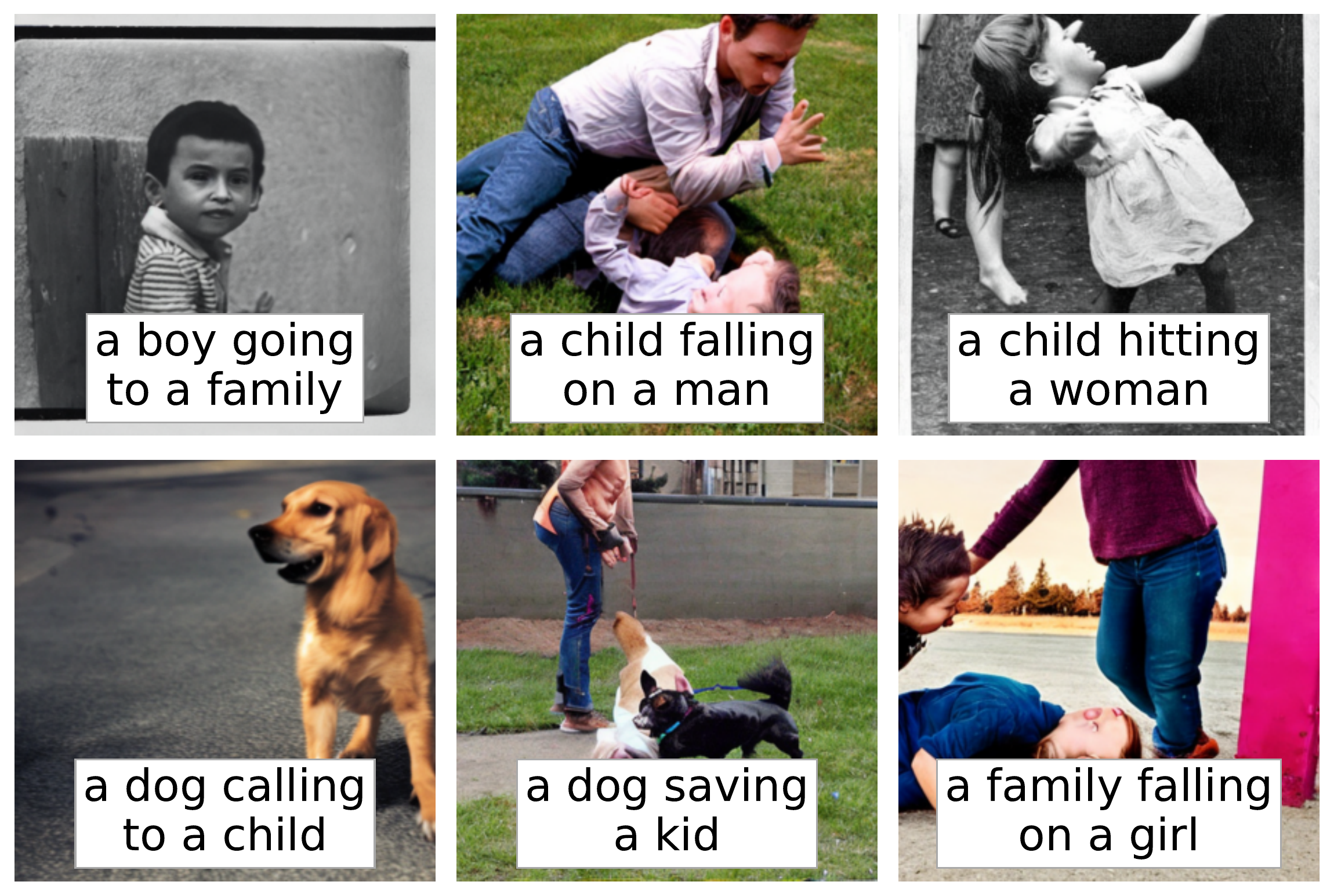}
\caption{\textbf{Failed Low-Frequency Generations that Support the \texttt{Forward} Hypothesis.} These examples appeared at most three times in the training data, sampled from the \textcolor{blue}{blue rectangle} in \cref{fig:exp1-chart}.
In five of six examples, only the subject or the object is present, not both. }\label{fig:exp1-samples-lf}
\end{figure}

\begin{figure}[ht!]
    \centering
    \includegraphics[width=0.95\linewidth]{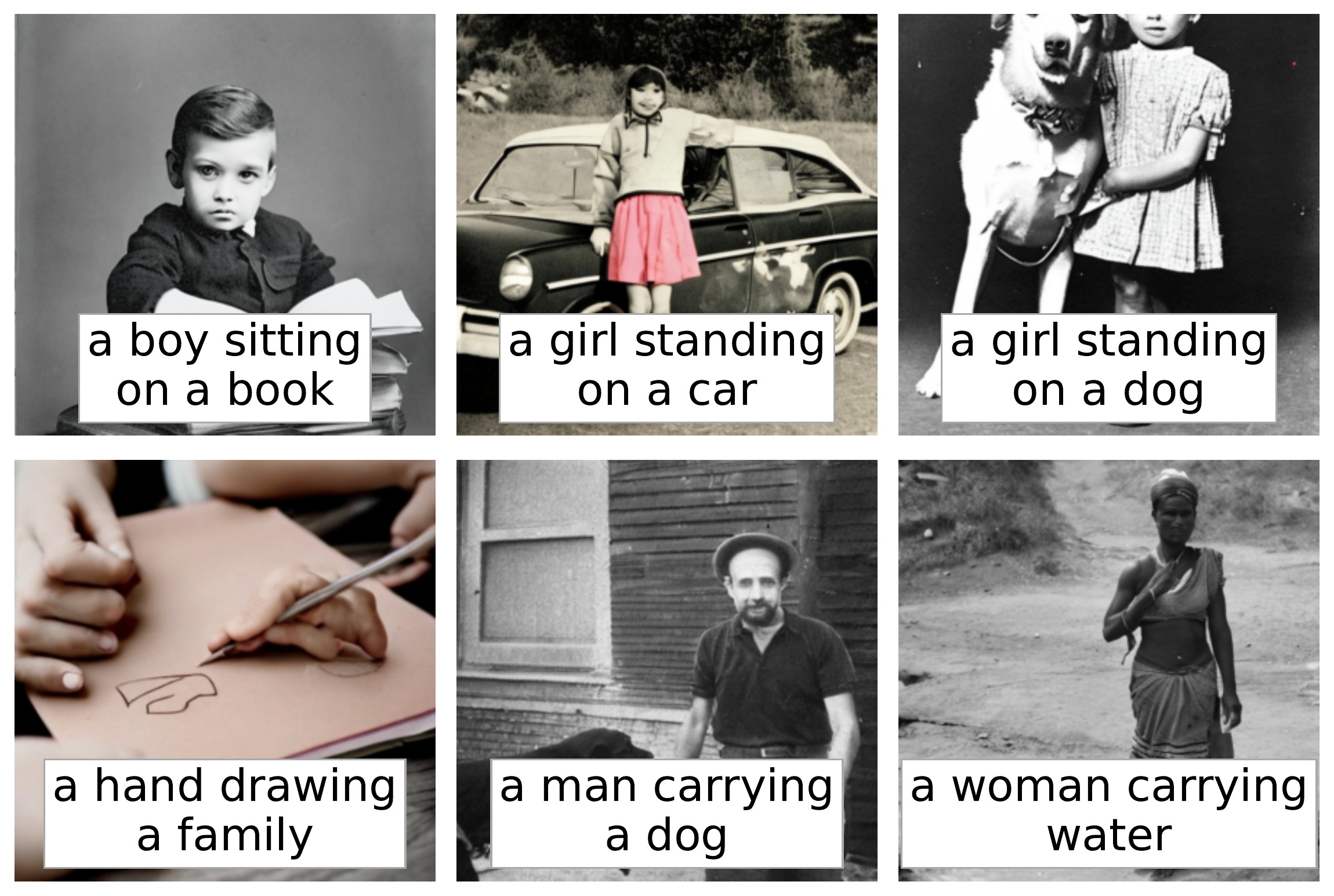}
\caption{\textbf{Failed High-Frequency Generations that Run Against the \texttt{Forward} Hypothesis.} These examples were all present in the training data with high frequency, sampled from the \textcolor{red}{red rectangle} in \cref{fig:exp1-chart}. In all six examples, the subject is present; In four of six, the object is present; In two of six, the model appropriately generates the relationship between the entities. }\label{fig:exp1-samples-hf}
\end{figure}

Secondly, the text-to-image model appears to fail in different ways at low versus high frequency. In a sample of low-frequency samples, highlighted by the \textcolor{blue}{blue rectangle} in \cref{fig:exp1-chart} and shown in \cref{fig:exp1-samples-lf}, either the subject or the object is generated but not both. Failed high-frequency, highlighted by the \textcolor{red}{red rectangle} in \cref{fig:exp1-chart} and shown in \cref{fig:exp1-samples-hf} differs. The subjects and objects are generated, but the verb is not properly shown. While failing, the high-frequency samples are closer to being aligned with the prompt. They appear to express a more common relation/verb than the one requested by the prompt, again pointing towards a training prior. 

\begin{figure}[ht!]
    \centering
    \includegraphics[width=0.95\linewidth]{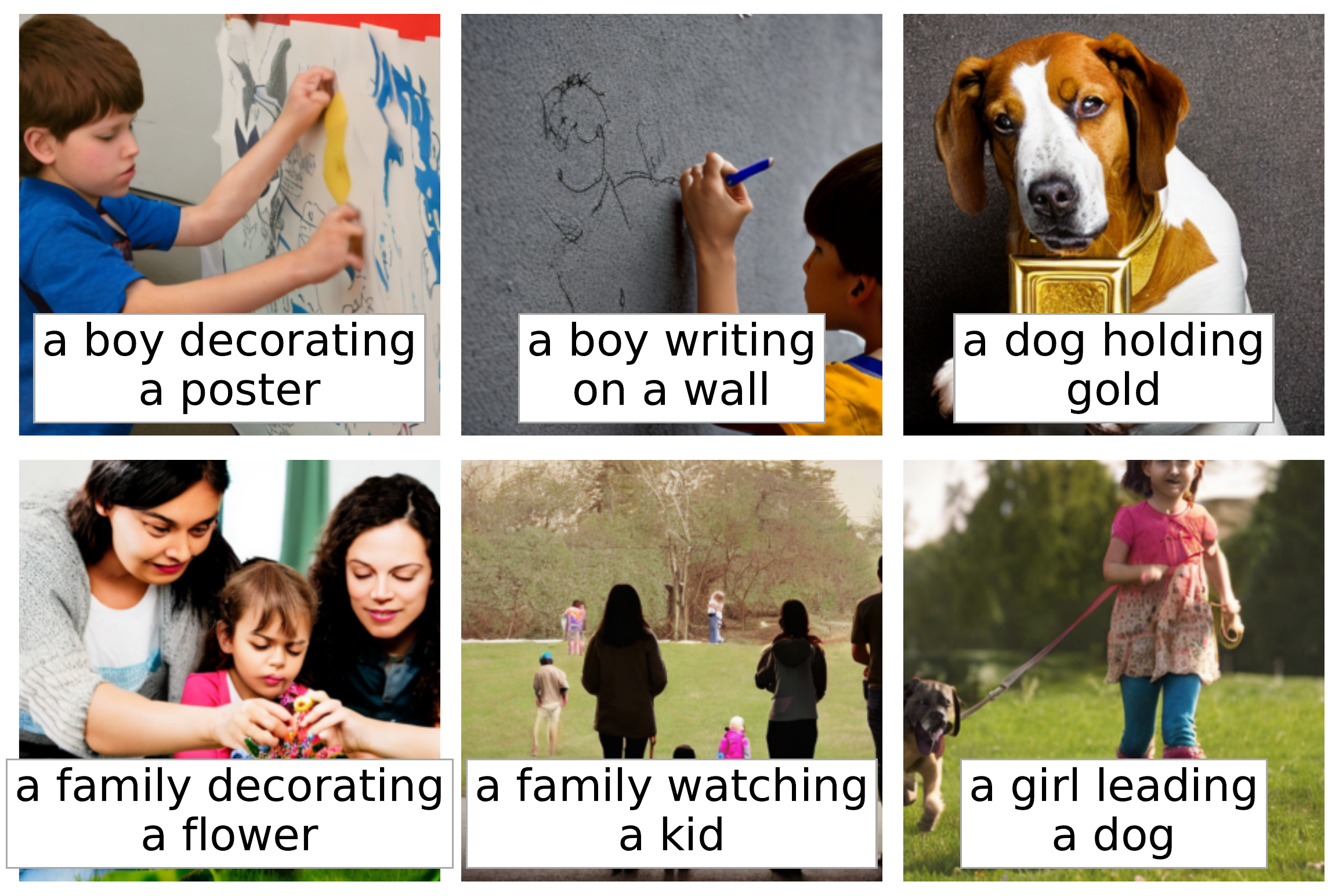}
\caption{\textbf{Successful Low-Frequency Generations that Run Against the \texttt{Forward} Hypothesis.} These relations (from the \textcolor{orange}{orange rectangle} in \cref{fig:exp1-chart}) were present at most 10 times in the training data.}\label{fig:exp1-samples-success-lf}
\end{figure}

\subsection{\texttt{Backward:} Increased OVS Decreases Alignment}\label{sec:exp2}
We established the frequency of a relation, say \dogball, improves the generation performance, given that the flipped relation, i.e., \balldog, was unseen. Now, we look at the opposite case. We generate images for unseen triads, say \balldog. Here, the inverse triads (\dogball) are seen to varying degrees. The question this section addresses is how does the text-to-image model behave when it is forced to fight a training prior? I.e., this section evaluates our \texttt{backward} hypothesis.

\begin{table}[ht!]
\centering
\begin{tabular}{lrr}
\toprule
&  \multicolumn{2}{c}{OVS Isolated \cref{sec:exp2}} \\
Term & Effect & \textit{p}-value \\
\midrule
OVS &  -0.06  &  0.38  \\
Sxx      &  0.07 & 0.01    \\
xVx      &   {0.13} & 0.00\\
xxO      &   { 0.19} & 0.00 \\ 
Oxx      &   0.10 & 0.01\\
xxS      &    {-0.19} & 0.00\\
\bottomrule
\end{tabular}
\caption{\textbf{Regression Coefficients} for Predicting Alignment (when SVO count is 0).
}
\label{tab:exp2}
\end{table}

\subsubsection{Results}
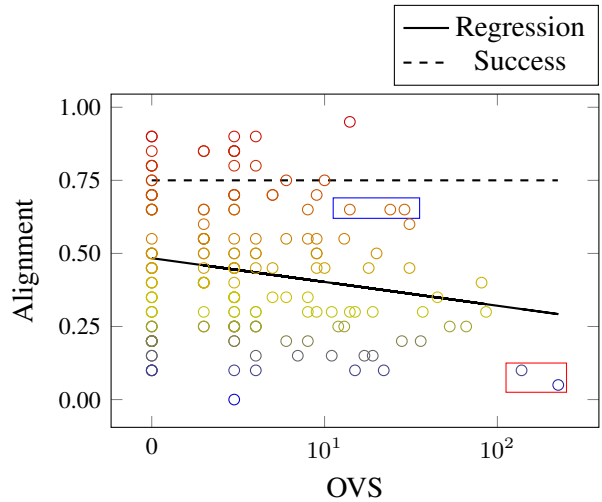
\begin{figure}[ht!]
         \centering 
\begin{tikzpicture}
\begin{axis}[
height=6cm,
width=8cm,
ylabel=Alignment,
legend style={at={(1.0,1.025)},anchor=south east},
xlabel=OVS,
ytick={0.00,0.25,0.50,0.75,1.00},
yticklabels={0.00,0.25,0.50,0.75,1.00},
xtick={0,1,2,3,4},
xticklabels={0,$10^1$,$10^2$,$10^3$},
]

\addplot [no markers, thick, black, table/x=InverseFrequencyLog] table [
y={create col/linear regression={x=InverseFrequencyLog,y=Alignment}}, 
] {tikz/exp2scatter.dat};
\addlegendentry{Regression};
\addplot[mark=none, thick, dashed, black] coordinates {(0,0.75) (2.35, 0.75)};
\addlegendentry{Success};
\draw (axis cs:1.05,0.62) rectangle (1.55, 0.69) [blue];
\draw (axis cs:2.05,0.025) rectangle (2.4, 0.125) [red];
\addplot [scatter, table/y=Alignment,
table/x=InverseFrequencyLog,
mark=o,only marks, blue] table {tikz/exp2scatter.dat};
\end{axis}
\end{tikzpicture}
         \caption{\textbf{Inverse Frequency Degrades Performance.}  The results weakly support the \texttt{backward} hypothesis that increased OVS decreases performance. We show the samples \textcolor{blue}{middle} in \cref{fig:exp2-samples} and   \textcolor{red}{right} in \cref{fig:exp2-samples-hf}.}
         \label{fig:exp2-chart}
     \end{figure}

Our results are shown in \cref{tab:exp2} and \cref{fig:exp2-chart}. It is worth highlighting, first, that alignment overall is poor. That is, when the model generates an unseen relation, the generated image is rarely judged to reflect the prompt. Moreover, the frequency of the flipped relation (OVS) does not have a significant effect on alignment. There is a slight (insignificant) negative effect for OVS, but this drop appears to be better explained by xxS: The more common the subject specifically is seen in the object role, the worse the generation.  For example, that \balldog\  is generated poorly seems more because a ``ball'' is commonly seen as the object (patient), not because the inverse triad as a whole (\dogball) is common.

The individual terms xVx and xxO positively impact the alignment: The more often words are used in a given slot, the better the generation. This result is intuitive, especially
given that we filter the data such that SVO could not impact the alignment. 

\begin{figure}[ht!]
    \centering
    \includegraphics[width=0.95\linewidth]{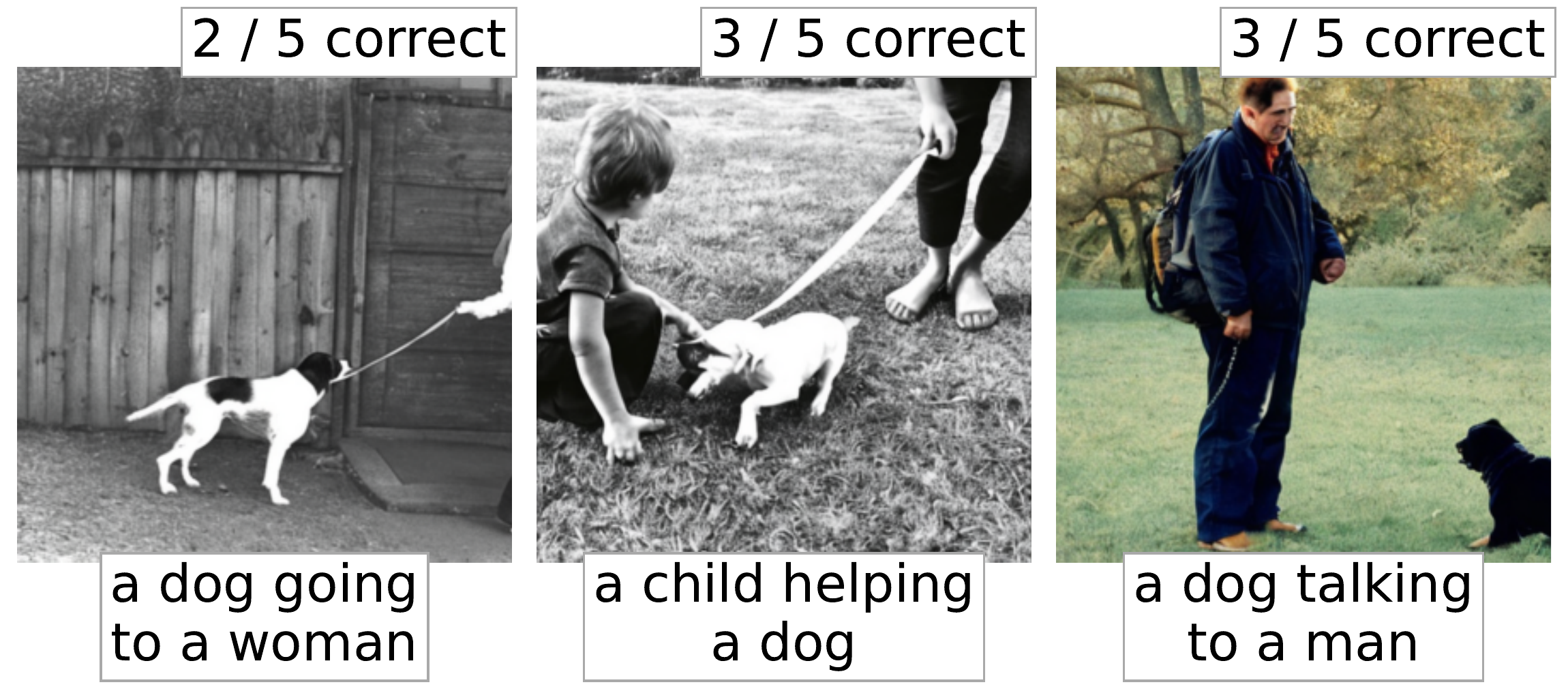}
\caption{\textbf{Examples Receive Ambiguous Ratings Decrease as OVS Increases.} Here we look at a set of examples highlighted in \cref{fig:exp2-chart} by the \textcolor{blue}{blue rectangle}. The average scores are below  success (0.75), but some raters considered the generations successful.}\label{fig:exp2-samples}
\end{figure}
\begin{figure}[ht!]
\centering
\includegraphics[width=0.69\linewidth]{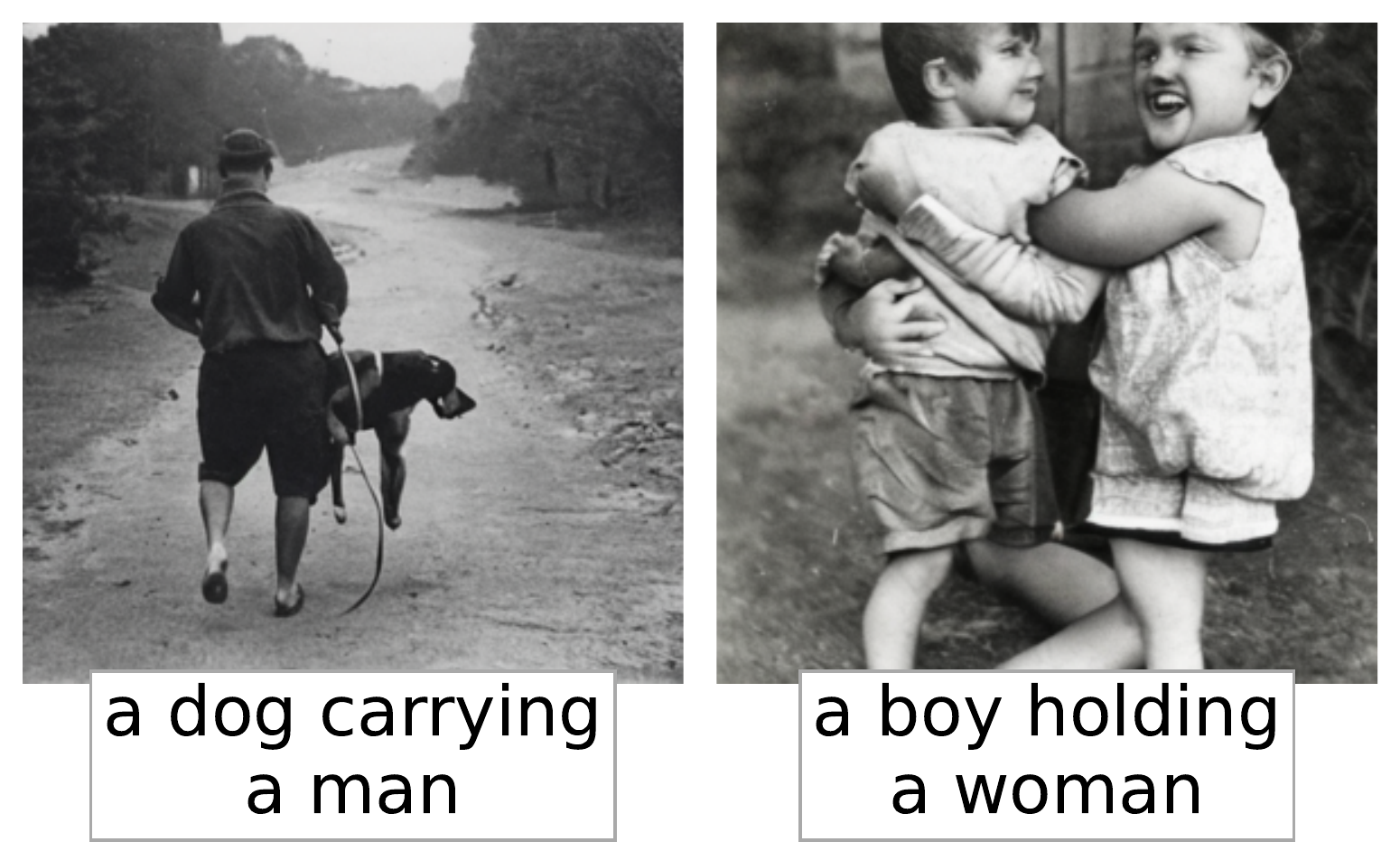}
\caption{\textbf{Examples Receive Low Ratings when OVS is High.} Here we look at a set of examples highlighted in \cref{fig:exp2-chart} by the \textcolor{red}{red rectangle}.}\label{fig:exp2-samples-hf}
\end{figure}

\subsection{Interaction between \texttt{Forward} and \texttt{Backward} Hypotheses}\label{sec:exp3}
The above results focused on cases when either SVO or OVS was zero--i.e., prompts for relations that are only attested in one direction. In this section, we focus our analysis on cases where the counts of SVO and OVS are both nonzero.

\subsubsection{Results}
The results, shown in \cref{tab:exp3}, largely support both hypotheses, but there are  differences between default and flipped examples.\footnote{When pooling all the examples (not shown) -- which introduces dependencies between the default and flipped examples -- the results are effectively dampened (slightly lowering the high effects we see for the default examples.)} Again, default cases are prompts where (say, \dogball) the relation is more common than the inverse relation (e.g., SVO $>$ OVS). Flipped is the opposite (say, \balldog): OVS $>$ SVO.

For default examples, e.g., \dogball, effects for both SVO ($+0.25$) and OVS ($-0.40$) are large. Again, as in \cref{sec:exp1}, Oxx ($-0.13$) and xxS ($-0.18$) also have a negative impact. For intuition, consider a specific example: \dogball. The more that ``ball'' has been seen in the subject role (and the more ``dog'' has been seen in the object role) \textit{in any context} the worse the alignment. 

For the flipped examples, e.g., \balldog, the effects are weaker. We expected a stronger result here because for these examples the model had to directly fight against strong priors. The overall lower alignment scores for this partition possibly impacted the results (see Appendix \cref{app:tab:datastats}).
 The most notable result is that SVO ($+0.14$)  still impacts the generations positively and that OVS ($-0.11$) is directionally aligned with the \texttt{backward} hypothesis (though the result is not significant.) 
 
\subsection{Takeaways} 
\textbf{The frequency effects were stronger for default examples, but there is evidence for both hypotheses.} This was unexpected (especially for OVS) because flipped examples are the cases where the model has the strongest prior to fight against. Therefore, our original motivation about why the text-to-image model fails to generate some relations is \textit{not} straightforwardly true: We did not find strong evidence that \horseastro is poorly generated because \astrohorse is common. However, in sum, our results do suggest that training priors impact the alignments in a number of ways. In Appendix \cref{app:tab:exp0}, we show that the regression coefficients across the full dataset match the individual patterns: Both primary hypotheses are supported, and the individual effects directionally align with the triads.

\textbf{Individual effects: term frequency in a given role  impacts the alignment.} Though we did not foresee this result, the logic runs in tandem with our primary hypotheses: Across the board, some of the alignment is explained by the triad frequencies--SVO and OVS--and some by the individual effects--Sxx, xVx, xxO, Oxx, and xxS. In \cref{sec:exp1} and \cref{sec:exp2}, this makes sense as the triads are prevented from impacting the results. However, these same strong individual effects are found for default examples in \cref{sec:exp3}, suggesting that training priors on individual terms also impact generations. This mirrors some human results \citep{mahowald2022experimentally} which show that agent/patient roles can be predicted from the words alone (no ordering) in a large majority of naturally occurring sentences.

\textbf{Semantic similarity between infrequent and frequent relations may further explain some of the results.} There are cases where low-frequency (low SVO) prompts were well generated; See \cref{fig:exp1-samples-success-lf}. A possible explanation is that if a relation is similar to another relation with high frequency in the training data, then that first relation may be better generated: For example, if  $\langle \textrm{dog, bite, bone} \rangle$ is  high frequency, then perhaps  $\langle \textrm{hound, chew, stick}\rangle$ is better generated.  More simply, how do the statistics of ``dog'' impact the generations of ``hound''? We leave the exploration of this idea to future work.

\begin{table}[ht!]
\centering
\begin{tabular}{lrr|rr}
\toprule
& \multicolumn{2}{c|}{Default \cref{sec:exp3}} & \multicolumn{2}{c}{Flipped \cref{sec:exp3}} \\
Term &  Effect & \textit{p}-value  &  Effect & \textit{p}-value \\
\midrule
SVO     &     {0.25} &   0.00 &  0.14  & 0.04\\
OVS & {-0.40} & 0.00 &  {-0.11}  &  0.11  \\
Sxx      &   {0.12}   & 0.00 & 0.03 & 0.27     \\
xVx      &    -0.03   &  0.34  & -0.05  & 0.12\\
xxO      &    0.07  & 0.16 & 0.08  & 0.11 \\ 
Oxx      &   {-0.13}  &  0.00  & -0.06 & 0.01 \\
xxS      &   {-0.18}  & 0.00  & 0.04 & 0.36\\
\bottomrule
\end{tabular}
\caption{\textbf{Regression Coefficients} for Predicting Alignment from OVS and SVO (and other factors).} 
\label{tab:exp3}
\end{table}

\section{Related Work}\label{sec:related}

\paragraph{Prompt Challenge Sets} 
Often released with a new text-to-image model, there are a number of prompt challenge sets that look at the capabilities of text-to-image models. These challenge sets, like this work, look at how and where text-to-image models fall short over different types of prompts: For example, DrawBench \citep{saharia2022photorealistic} \footnote{\url{https://docs.google.com/spreadsheets/d/1y7nAbmR4FREi6npB1u-Bo3GFdwdOPYJc617rBOxIRHY/edit\#gid=0}} has prompts across a range of categories like ``Conflicting'' and ``Colors''. See some examples in \cref{fig:drawbench}. However, in contrast to our work, these prompts are constructed largely by intuition rather than sourced by parsing the training data of the model. Another benchmark \citep{yu2022scaling}\footnote{\url{https://github.com/google-research/parti/blob/main/PartiPrompts.tsv}} test a broader set of phenomena beyond the scope of this work. Finally, WinoGround \citep{thrush2022winoground}, gets at the same ideas we are interested in: how and whether a vision+language model captures small differences in captions.

\paragraph{Improving Diffusion Models} Establishing how good (or not) current text-to-image models are at generating different types of relations provides context for future (and ongoing) work. The success of RLHF (Reinforcement Learning with Human Feedback) in improving large language models \citep{Ziegler2019,NEURIPS2020_1f89885d,Ouyang2022,Gao2022,bai2022training} suggests that this type of training could also improve diffusion models. Already, we see some success: \citet{hao2022optimizing} used reinforcement learning to suggest additional tags for prompts to improve the generations (like ``high definition'', ``8k''); \citet{lee2023aligning} use RLHF to improve stablediffusion (although the weights are not public.) There is a growing literature in this area that experiment with guiding the diffusion process \citep{yang2022diffusionbased,feng2023trainingfree,po2023compositional,wang2023diffusionbased,liu2023compositional,chen2023trainingfree}.

\section{Discussion}
The marked success of recent text-to-image models \citep{ramesh2022hierarchical,saharia2022photorealistic} may undermine prior intuitions about the compositional capacity of neural network models. This raises questions about what type of mechanism supports such generalization--are neural models using something that resembles abstract combinatorial structure, or some form of clever mix-and-match strategy? In this work, we found that the  counts of the underlying relations in prompts significantly impact the generations of a stablediffusion 2.1. Although not definitive, such results suggest the latter strategy (mix-and-match). Despite the important caveat that our results only reflect the behavior of one model,
they still generate several interesting points worthy of discussion and further exploration.

\textbf{Implications for compositionality of neural networks.} The primary question motivating this work is: can neural networks reason compositionally, and if so, how? The results we present only truly weigh on the second part of this question--i.e., our results strongly suggest that models employ some form of content-specific, data-dependent representations rather than using abstract systematic processes. A similar observation was suggested by \cite{chaabouni-etal-2020-compositionality}. This result is both surprising and obvious. It is obvious because, in the history of debates about neural networks and compositionality, it has always been asserted that they would use such idiosyncratic strategies to solve tasks. However, it is surprising because such methods appear to bring us much further in producing behavior that has all the signatures of compositionality than previously thought. That is--behavior of the type we currently see in image-, language-, and code-generation models were once thought to, in and of itself, be a diagnostic for abstract combinatorial structure. Increasingly, it seems to be the case that such structure is not a prerequisite for compositional (appearing) behavior \cite{chaabouni-etal-2020-compositionality}. In our analysis, for example, there exist examples of low-frequency triads with high alignment scores. A very important question for future work is to revisit what, exactly, we seek when we seek models which are compositional: are we placing requirements on the behavior, the mechanism, or some combination of the two? 

\section*{Limitations}\label{sec:limitations}
\textbf{We focus on a particular type of construction}: subject, verb, object. Furthermore, we do not nest these types of prompts, e.g., ``the dog chased the man holding the child.'' We also do not focus on adjectives, like color. These types of ideas have been touched upon by previous work \citep{saharia2022photorealistic}. Many sentences in the dataset are not typical sentences and end up being unattested to in our dataset counts. Instead, many are descriptive fragments like ``Suki Round Coffee Table'' or ``columbia valley wine collections chateau the best cabernet sauvignon 20 the wine snob picks.'' Our counts do not capture how often terms are used in these types of sentences.

\textbf{Assumptions of our experimental design}: (1) The relations pertain to the meaning of the sentences. (2) The sentences pertain to the content of the images. (3) The true statistics of relations in the dataset are proportional to the statistics we parsed.

We investigate each of these concerns in \cref{app:sec:collection}. Though the counts we provide are estimates, for a sample of triads, a deeper dive through the dataset does not yield significantly different counts. Importantly: Unseen triads remained unseen and seen triads had similar counts as originally found. Finally, there are a couple of notes that we consider out of scope and leave to follow-up work: First, our focus is on the training dataset of the diffusion model, not the text model. Second, our experiments assume that the counts of triads is the causal factor leading to deltas in alignment scores. Ultimately, the reason that certain triads are frequent and others are not will impact more than the frequency counts alone. 

\textbf{Our experimental design requires access to the training data}, and then we search and parse the training data. First, this is a slow process. Second, this precludes testing commercially available, but closed, models like DALLE under our framework.

\section*{Ethics Statement}
Our work aims to help us understand how text-to-image models work, and enable us to better understand the types of things models are currently capable of. Though the triads we find are generic, this paper makes clear that this type of text-to-image model will tend to reflect the relations seen in the wild.

\bibliography{emnlp}
\bibliographystyle{abbrvnat}

\appendix
\section{Dataset Collection}\label{app:sec:collection}
\subsection{Details}
We parse sentences from LAION with aesthetic scores above 4.5 because this is the training data of stable diffusion 2.1. (Stablediffusion 2.1 also filtered out images that have a probablity of being unsafe above 0.1. In our process, we did not do this additional filtering.)  We access the dataset via HuggingFace at \url{https://huggingface.co/datasets/ChristophSchuhmann/improved_aesthetics_4.5plus} which has 1372 million rows. We successfully parse 134 million sentences with 50 million unique subject-verb-object triads (with 1,687,675 unique subjects, 669,858 unique verbs, and 1,553,343 unique objects.) As detailed below, we looked for common individual terms which appeared in many contexts, significantly reducing the pool of sentences.

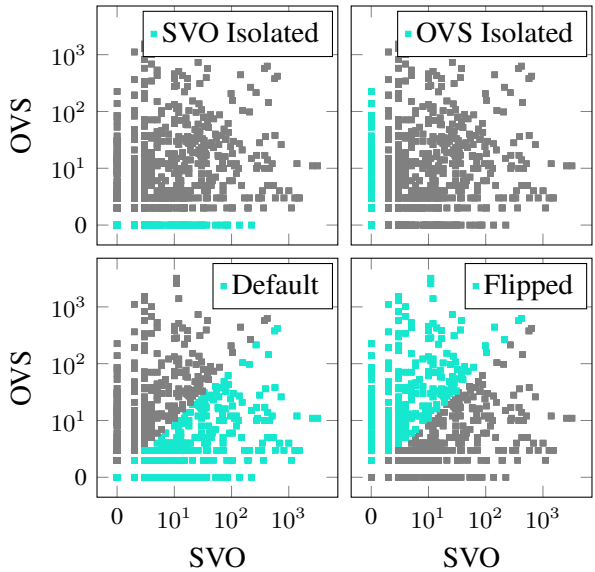
\begin{figure}[ht!]
\centering 
\begin{tikzpicture}
\begin{groupplot}[group 
style={columns=3},
height=4.75cm,
width=4.75cm,
xtick={0,1,2,3,4},
xticklabels={0,$10^1$,$10^2$,$10^3$},
ytick={0,1,2,3,4},
yticklabels={0,$10^1$,$10^2$,$10^3$},
xlabel=SVO,
ylabel=OVS,
group style={
    group size=2 by 2,
    horizontal sep=5pt,
    vertical sep=5pt,
    x descriptions at=edge bottom,
    y descriptions at=edge left},
]
\nextgroupplot [
]
\addplot [scatter, 
table/y=OVS,
table/x=SVO,
table/meta=Isolated,
point meta=explicit,
mark=square*, mark size=1pt,
only marks,
scatter/classes={
0={gray},
1={C7}
}
] table {tikzdataset/dataset_svo_isolated.dat};
\legend{,SVO Isolated};

\nextgroupplot [
]
\addplot [scatter, 
table/y=OVS,
table/x=SVO,
table/meta=Isolated,
point meta=explicit,
mark=square*, mark size=1pt,
only marks,
scatter/classes={
0={gray},
1={C7}
}
] table {tikzdataset/dataset_ovs_isolated.dat};

\legend{,OVS Isolated};

\nextgroupplot [
]
\addplot [scatter, 
table/y=OVS,
table/x=SVO,
table/meta=Default,
point meta=explicit,
mark=square*, mark size=1pt,
only marks,
scatter/classes={
0={gray},
1={C7}
}
] table {tikzdataset/dataset_default.dat};

\legend{,Default};

\nextgroupplot [
]
\addplot [scatter, 
table/y=OVS,
table/x=SVO,
table/meta=Flipped,
point meta=explicit,
mark=square*, mark size=1pt,
only marks,
scatter/classes={
0={gray},
1={C7}
}
] table {tikzdataset/dataset_flipped.dat};
\legend{,Flipped};
\end{groupplot}
\end{tikzpicture}
\caption{These plots show how the full dataset is split up the various view across \cref{sec:exp1}, \cref{sec:exp2}, \cref{sec:exp3}.}\label{app:fig:partitions}
\end{figure}

\cref{app:fig:partitions} visually show how the dataset is split across the various experiments across \cref{sec:exp1}, \cref{sec:exp2}, \cref{sec:exp3}. \cref{app:tab:datastats} shows the average alignment scores.

\begin{table*}[ht!]
\centering
\begin{tabular}{lrrrr|r}
\toprule
    & SVO Iso. & OVS Iso. & Default & Flipped & All\\
\midrule
count & 755 & 835 & 1635 & 1635 & 4323 \\
mean  &   0.56 & 0.44 & 0.50 & 0.46 & 0.49 \\
std   &   0.34 & 0.31& 0.33 & 0.33 & 0.33 \\
min   &   0.00 & 0.00 & 0.00 & 0.00 & 0.00 \\
25\%  &    0.25 & 0.25 & 0.25 & 0.25 & 0.25\\
50\%   &   0.75 & 0.25 & 0.50 & 0.25 & 0.5 \\
75\%   &   0.75 & 0.75 & 0.75 & 0.75 & 0.75 \\
max   &   1.00 & 1.00 &  1.00 & 1.00 & 1.00 \\
\bottomrule
\end{tabular}
\caption{\textbf{Dataset Partition Stats} for Predicting Alignment from OVS and SVO (and other factors). There are 769 total unique prompts/relations. The difference in mean and median (50\%) values between the first and third vs second and fourth highlights the relative difficulty of the second pair.} 
\label{app:tab:datastats}
\end{table*}

\subsection{Evaluating Our Assumptions}\label{app:sec:assumptions}
\paragraph{Seen Triads are Seen}
We randomly sampled the following seen tuples: \cref{app:fig:assumptions1}. (1) We search the training dataset for sentences that contain all three words. (2) We re-parse these found sentences to get a parse-rate for each triad. (3) We manually label $2 * 4 * 25 = 200$ sentences (25 per triad) (parsed and not) looking for a parse-intent-possible. Here, we're informally (liberally) looking to find sentences where the searched terms could plausibly be the subject, verb and object. The sampled sentences: \url{https://docs.google.com/spreadsheets/d/1tViq4rYLXvjzA4vqhEFB-TqXvo4PzRt2ezAlEZrOvVI/edit?usp=sharing}. (4) We estimate the total number examples for each of the triads by summing the number of parsed (and not parsed) examples by the parse-intent-possible, respectively. We find that the estimates are in general larger than the number of sentences we originally found, but not beyond 2x. We do not compute estimates of but instead use the raw counts from the collected data.

In the sampled triads, the OVS counts were low. We found no examples of the flipped terms when the sentence was parsed (in part because they are much more rare.) The number of actual flipped examples amond the very few parsed examples we did find was extremely high (near perfect).

\begin{figure*}
\centering
    \begin{tabular}{lrr|rr}
\toprule
Triad & SVO & OVS & Estimated SVO & Estimated OVS \\
\midrule
$\langle \textrm{girl, play, dog} \rangle$ & 1389 & 11 & 3228 & 11\\
$\langle \textrm{woman, walk, dog} \rangle$ & 1444 & 10 & 2652 & 8\\
$\langle \textrm{woman, carry, child} \rangle$ & 1536 & 2 & 1076 & 5 \\
$\langle \textrm{woman, play, dog} \rangle$ & 3207 & 2 & 7444 & 3\\
\bottomrule
\end{tabular}
\caption{\textbf{Seen Triads are Seen.}}
\label{app:fig:assumptions1}
\end{figure*}

\paragraph{Unseen Triads are Unseen}
Like with the seen triads, we sampled the following unseen tuples (\cref{app:fig:assumptions2}) and follow the same procedure as above. Somewhat surprisingly, we found no examples of the unseen tuples, although ultimately, because we are making use of sampling, it is quite possible that there are some examples. To be clear: There were still no cases of the unseen triads being parsed and no sampled examples of these triads.  Of note: The rate with which even one of the three words took a subject, object, or verb role was below 3\% of the sentences. In comparison, for the seen triads, the individual terms matched on average for subjects 23\%, for verbs 17\%, and 24\% for objects. (The sampled sentences are in the same document as above, \url{https://docs.google.com/spreadsheets/d/1tViq4rYLXvjzA4vqhEFB-TqXvo4PzRt2ezAlEZrOvVI/edit?usp=sharing}.)
\begin{figure}
    \centering
\begin{tabular}{lrr}
\toprule
Triad & SVO & OVS \\
\midrule
$\langle \textrm{girl, move, dog} \rangle$ & 0 & 0 \\
$\langle \textrm{girl, fall, family} \rangle$ & 0 & 0 \\
$\langle \textrm{man, fall, family} \rangle$ & 0 & 0 \\
$\langle \textrm{man, hit, kid} \rangle$ & 0 & 0 \\
\bottomrule
\end{tabular}
\caption{\textbf{Unseen Triads are Unseen.}}
\label{app:fig:assumptions2}
\end{figure}

\paragraph{Triads Pertain to Sentence Meaning; Images and Sentences Align} We take 100 parsed and 100 not-parsed sentences sampled from the training data: \url{https://docs.google.com/spreadsheets/d/1hIj-PvQsHM6WnRjRVhK1OKp3fOVp2wwjVjdKLRZwmIs/edit?usp=sharing}. (Note: The distribution of these sentences differs from the high-frequency triads and terms, but provides a baseline window into the parsing process.) We find:
\begin{enumerate}
    \item 5 of 100 not-parsed are false negatives (failed to be parsed when they could have been.)
    \item 36 of 100 parsed examples are false positives (should not have been parsed)
    \item 39 of 100 parsed examples had the clear potential for being successfully parsed (called possible parses below).
    \item 36 of the 39 possible parses, the parsed triad was the correct triad.
    \item 35 of the 39 possible parses, the parsed triad captured the meaning of the sentence.
    \item 34 of the 39 possible parses, the sentence captured the meaning of the photo.
\end{enumerate}

In sum, the process is noisy, but we are finding reasonable success: For most examples that are possible parses, the parsed triad matches the meaning of the sentence and the paired photo matches the sentence. Detailed in the subsection below, we select individual terms that are common, with the aim of handling some of the noisy parses and finding some safety in the higher counts.

\subsection{Protocol for SVO Triads}
We will make the source code for this process public in future release of this paper after peer review.
\begin{enumerate}
\item Parse dataset for SVO triads using spacy small english parser.
\item Filter for alpha characters only.
\item Get the number of instances each triad appears. For example, we build an index: S → Count[triads S appears in].
\item Threshold such that each individual word (S, V, O) has been seen N = 5000 times.
\item Find the intersection of S and O.
\item Choose the top 100 for each S, V, O (using the intersection for S and O).
\item Build index (triad → count)
\item Determine attestation for each triad in a crossed dataset. Attestation is one of (FF, TT, FT, TF) where FF is neither direction of the triad has been seen, TT is both directions of the triad are seen, FT is the backwards direction has been seen, and TF is the forward direction has been seen.
\item For each split of attestation (FF, TT, FT, TF) sample 15 random triads per verb.
\item Manually label each triad for whether it is drawable in both directions (SVO, OVS). This step is subjective. This step could be automated using online workers. Because the drawable images are relatively sparse, we found it overall easier to do the labelling ourselves and avoid the overhead.
\item Using this approach found a limited number of high-frequency tuples. We expanded the pool of candidate triads by bootstrapping off the successful cases and finding more examples of high-frequency tuples.
\end{enumerate}

\section{Labelling Task Interface}\label{app:sec:label}
\begin{figure*}[ht!]
         \centering 
\includegraphics[width=16cm,keepaspectratio]{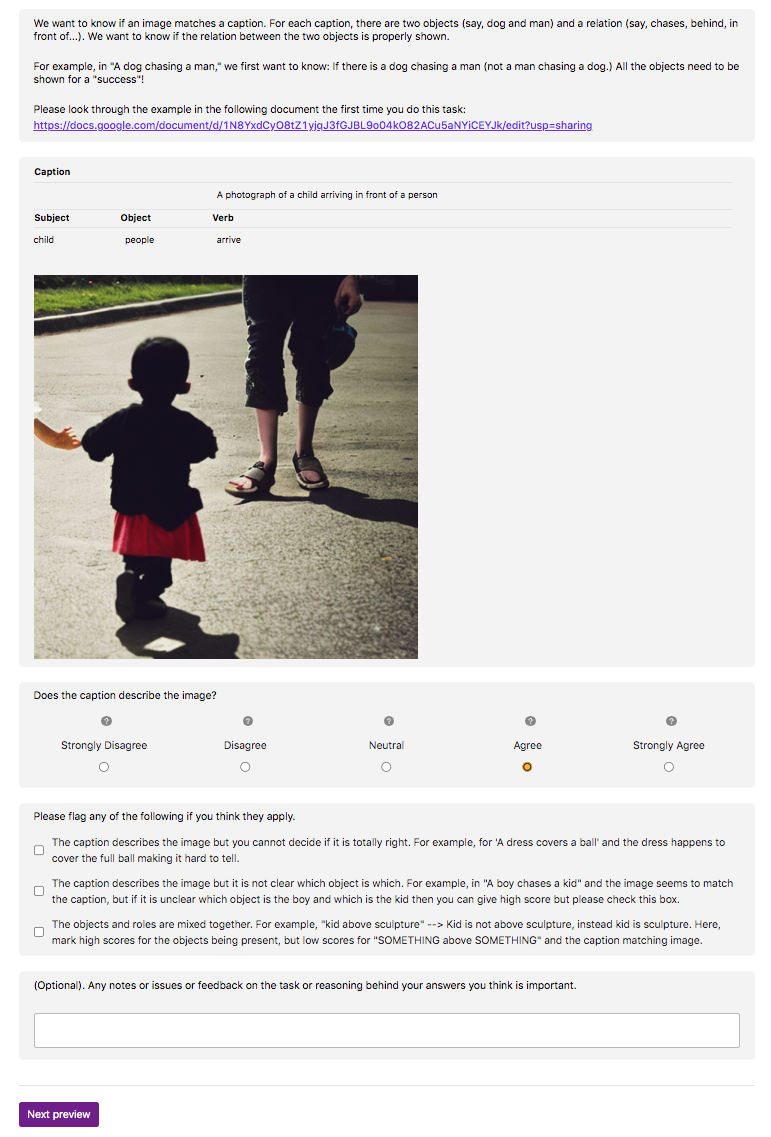}
         \caption{\textbf{Interface for Human Preferences.}  } 
         \label{figure:interface}
     \end{figure*}

\cref{figure:interface} is the interface we used to collect human judgements on the text/caption alignment. We provided examples to the SurgeAI workers are the following link:
\url{https://docs.google.com/document/d/1N8YxdCyO8tZ1yjqJ3fGJBL9o04kO82ACu5aNYiCEYJk/edit?usp=sharing}.

\section{Additional Results}
\subsection{Alternate Regressions}
The results across all the partitions combined are in \cref{app:tab:exp0}.
The results from \cref{tab:exp2} are reproduced but with random controls in \cref{app:tab:exp2:controls} and averaged scores in \cref{app:tab:exp2:average}.

\begin{table}[ht!]
\centering
\begin{tabular}{lrr}
\toprule
Term &  Effect & \textit{p}-value \\
\midrule
SVO     &     {0.25} &   0.00\\
OVS & {-0.20} &  0.00 \\
Sxx      &   {0.04}   & 0.001\\
xVx      &    0.01   &  0.79 \\
xxO      &     0.13  & 0.00\\
Oxx      &   {-0.08}  &  0.00\\
xxS      &   {-0.08}  & 0.001\\
\bottomrule
\end{tabular}
\caption{\textbf{Regression Coefficients for All Partitions Combined} for Predicting Alignment from OVS and SVO (and other factors).} 
\label{app:tab:exp0}
\end{table}

\begin{table}[ht!]
\centering
\begin{tabular}{lrr|rr}
\toprule
& \multicolumn{2}{c|}{Default} & \multicolumn{2}{c}{Flipped} \\
Term &  Effect & \textit{t}-value  &  Effect & \textit{t}-value \\
\midrule
SVO     &     {0.24} &   3.79 &  0.12  & 2.00\\
OVS & {-0.41} &  -6.59 &  -0.14  &  -2.15  \\
Sxx      &   {0.11}   & 4.93 & 0.03 & 1.16   \\
xVx      &    -0.02   &  -0.86  & -0.06  & -2.06 \\
xxO      &    0.06  & 1.32 & 0.08 & 1.63 \\ 
Oxx      &   {-0.13}  &  -4.95  & -0.06 & -2.66 \\
xxS      &   {-0.19}  & -3.96  & 0.03 & 0.65\\
\bottomrule
\end{tabular}
\caption{\textbf{Regression Coefficients with Raters as Random Controls} for Predicting Alignment from OVS and SVO (and other factors).} 
\label{app:tab:exp2:controls}
\end{table}

\begin{table}[ht!]
\centering
\begin{tabular}{lrr|rr}
\toprule
& \multicolumn{2}{c|}{Default} & \multicolumn{2}{c}{Flipped} \\
Term &  Effect & \textit{p}-value  &  Effect & \textit{p}-value \\
\midrule
SVO     &     {0.25}  & 0.04 &  0.14  & 0.23\\
OVS     &    {-0.10}  & 0.00 & -0.11  &  0.35  \\
Sxx      &   {0.12}   & 0.01 & 0.03 & 0.52     \\
xVx      &    -0.03   & 0.58 & -0.05  & 0.36\\
xxO      &    0.08    & 0.42 & 0.08  & 0.35 \\ 
Oxx      &   {-0.06}  & 0.01 & -0.06 & 0.15 \\
xxS       &   {0.04}  & 0.03 & 0.04  & 0.60 \\
\bottomrule
\end{tabular}
\caption{\textbf{Regression Coefficients with Averaged Scores} for Predicting Alignment from OVS and SVO (and other factors).} 
\label{app:tab:exp2:average}
\end{table}

\subsection{Controls}\label{sec:control}
We re-run our experiments in \cref{sec:exp3} but present to the raters images from the flipped prompt. For example, in Control Default, we present the raters the prompt from \dogball\  and the image generated from \balldog. (In Control Flipped, we show the raters the prompt from \balldog\  and the image generated from \dogball.)
 
\textbf{Predicted Outcomes:} A positive correlation between SVO and the swapped images suggests that the increased count of the SVO lead the flipped prompt to still generate the relation described by \svo. A negative correlation between OVS and the swapped images suggests that the more frequent the OVS count is the better the generation of the flipped prompt leading to the flipped relation. Both of these are direct predictions of the \texttt{forward} and \texttt{backward} hypotheses we outline in \cref{sec:hyp}.

\textbf{Alternate Outcomes:} If the reverse of the above held--if SVO had a negative correlation or OVS had a positive correlation--this would provide evidence counter to our hypotheses.

\textbf{Results:} The results largely match the predicted outcomes: \cref{fig:control}. We see that the Control Default has a stronger effect on OVS whereas the Control Flipped has strong effects on both SVO and OVS. We don't take this evidence as bringing much more to bear on the problem than what we found in the original experiments.

\begin{table}[ht!]
\centering
\begin{tabular}{lrr|rr}
\toprule
& \multicolumn{2}{c|}{Control Default} & \multicolumn{2}{c}{Control Flipped} \\
Term &  Effect & \textit{p}-value  &  Effect & \textit{p}-value \\
\midrule
SVO     &    {0.04} &   0.53 & \textbf{0.20 }& 0.00\\
OVS & \textbf{-0.27} & 0.00 &  \textbf{-0.14}  &  0.02  \\
Sxx      &   \textbf{0.12}   & 0.00 & 0.01 & 0.70    \\
xVx      &    -0.08   &  0.02 & -0.00 & 0.94 \\
xxO      &    -{0.05}  & 0.27 & \textbf{0.14}  & 0.00 \\ 
Oxx      &   {-0.01}  &  0.62  & -0.05 & 0.02 \\
xxS      &   \textbf{-0.13}  & 0.01  & -0.07 & 0.10\\
\bottomrule
\end{tabular}
\caption{\textbf{Regression Coefficients for Control Experiments.} Significant ($p < 0.05$) effects over 0.1 are \textbf{bolded}. }
\label{fig:control}
\end{table}

\subsection{Prior Work}
We evaluate how well stablediffusion 2.1 generates images aligned with prompts from prior work. We report alignment scores on prior work in \cref{tbl:other-datasets}, \cref{app:tbl:other-datasets}, and \cref{app:tbl:other-datasets-cont}. Unlike in the main body of the work, the scores  are from 0 to 4. 

\begin{figure}
\centering
    \includegraphics[width=0.3\linewidth]{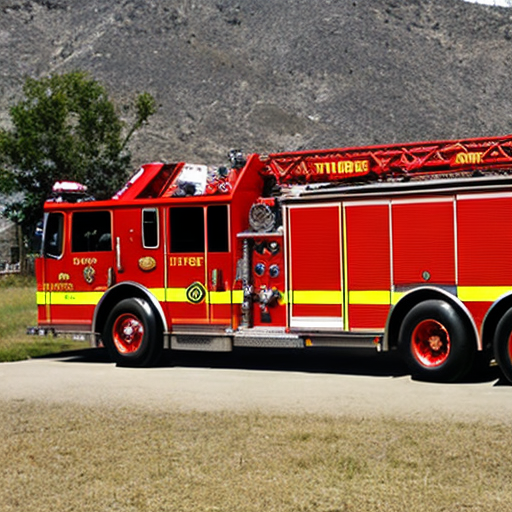}
    \includegraphics[width=0.3\linewidth]{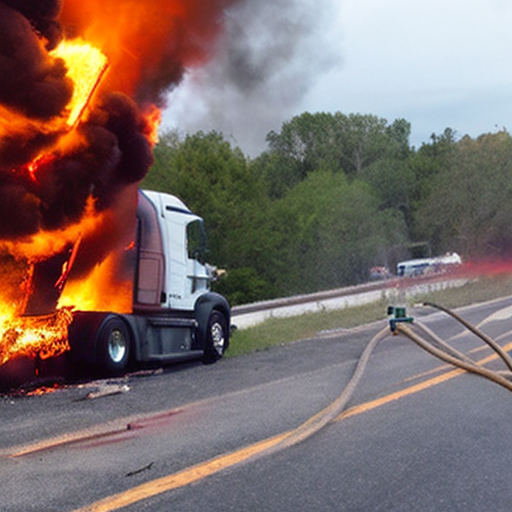}\\
    \includegraphics[width=0.3\linewidth]{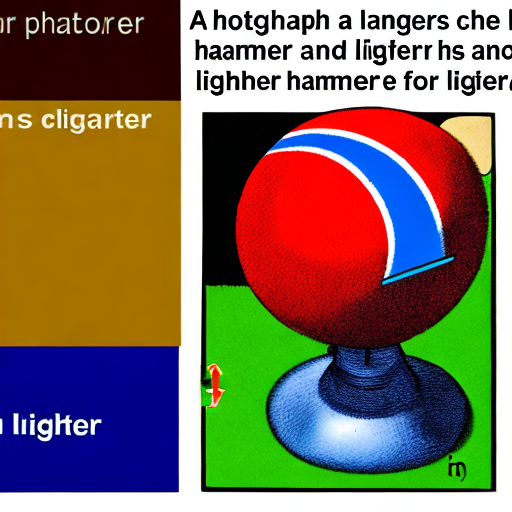}
    \includegraphics[width=0.3\linewidth]{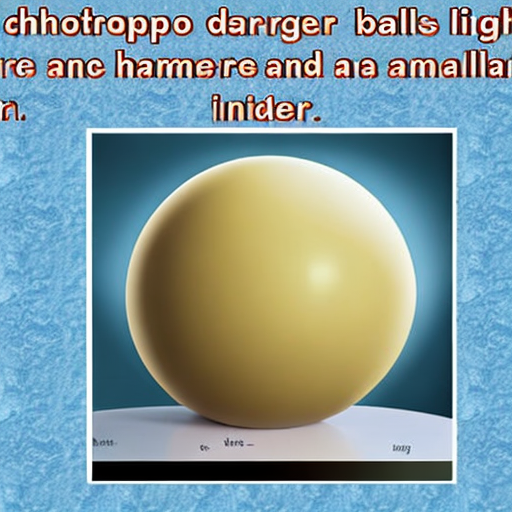}
\caption{\textbf{WinoGround example generations.} Top: ``There is a [fire] [truck]'' vs ``There is a [truck] [fire.]''; Bottom: ``a larger ball is lighter and a smaller one is heavier'' vs ``a larger ball is heavier and a smaller one is lighter''. The top examples belong to the Noun category; the bottom examples belong to Adjective-Weight. These examples were generated by stablediffusion 2.1.}
\label{fig:winoground}
\end{figure}

 Our results on Appendix \cref{tbl:other-datasets} suggest that WinoGround is also a difficult dataset for text-to-image models. (Vision-and-language models were unable to perform much beyond chance upon it when set to disambiguate paired captions). \cref{fig:winoground} shows some successes and failures. stablediffusion 2.1 was successful on the Noun-based prompts but failed on a descriptive prompt that required understanding the weight of different types of balls. (Presumably, the desired generations involved things like first a bowling ball and a beach ball, and then a bowling ball and a tennis ball.)

\begin{figure}
\centering
  \centering
    \includegraphics[width=0.3\linewidth]{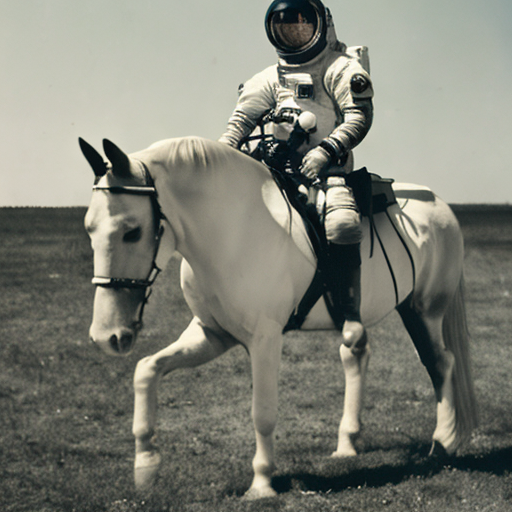}
    \includegraphics[width=0.3\linewidth]{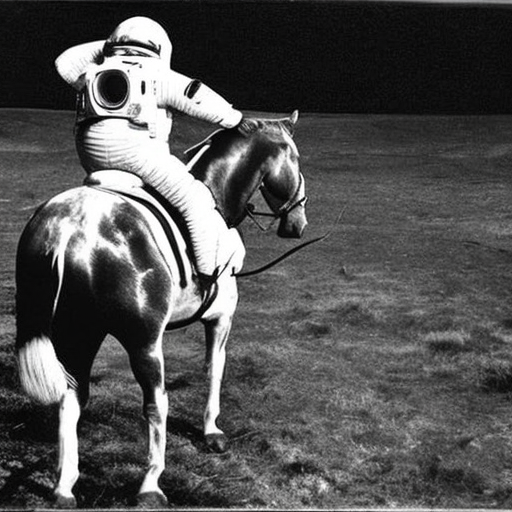}\\
    \includegraphics[width=0.3\linewidth]{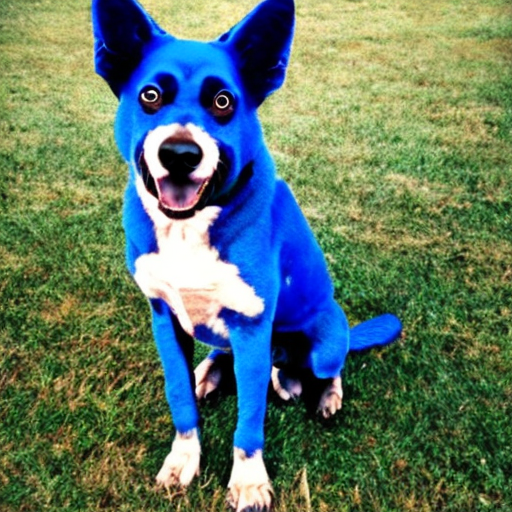}
    \includegraphics[width=0.3\linewidth]{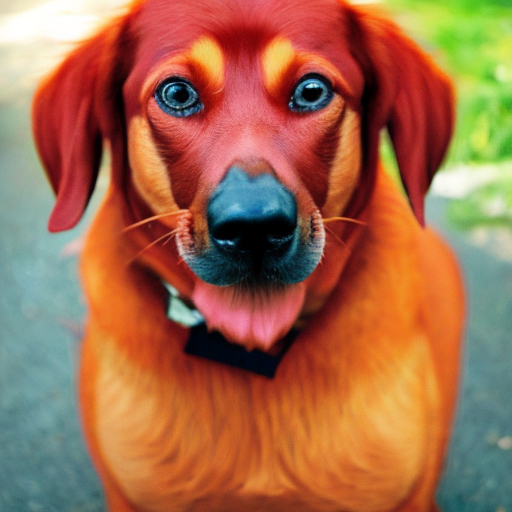} 
\caption{\textbf{DrawBench example generations.} Top left: ``horse riding an astronaut''; Top right: ``astronaut riding a horse''; Bot left: ``blue colored dog''; Bot right: ``red colored dog''. The first two images, belonging to the ``Conflicting'' category show how the model was biased towards the typical relation. The second two images show the model behaving more flexibly. These examples were generated by stablediffusion 2.1.}
\label{fig:drawbench}
\end{figure}

\begin{table*}[ht!]
\centering

\begin{tabular}{llrrrr}
\toprule
{} &             Category & Count &  Median &  0.25 &  0.75 \\
\midrule
\texttt{DrawBench} & Rare Words &      7 &     0.0 &   0.00 &   1.50 \\
          & Misspellings &     10 &     0.5 &   0.00 &   2.75 \\
          & Gary Marcus et al.  &     10 &     1.0 &   0.25 &   3.00 \\
          & Positional &     20 &     1.0 &   1.00 &   2.00 \\
          & Reddit &     38 &     1.0 &   1.00 &   3.00 \\
          & Text &     21 &     1.0 &   0.00 &   2.00 \\
          & Counting &     19 &     2.0 &   1.00 &   4.00 \\
          & DALL-E &     20 &     2.5 &   0.75 &   3.00 \\
          & Colors &     25 &     3.0 &   3.00 &   4.00 \\
          & Conflicting &     10 &     3.0 &   1.50 &   4.00 \\
          & Descriptions &     20 &     3.0 &   1.75 &   3.25 \\
\texttt{WinoGround}  & Both &     52 &     1.0 &    1.0 &    3.0 \\
           & Relation &    466 &     1.0 &    1.0 &    3.0 \\
           & Object &    282 &     2.0 &    1.0 &    3.0 \\
\bottomrule\\
\end{tabular}

\caption{\textbf{Alignment Scores on Prior Work}. The alignment columns reports the median and quartiles. A fine-grained breakdown of categories in WinoGround can be found in \cref{app:tbl:other-datasets}. The alignment scores are scaled from 0 to 4.}
\label{tbl:other-datasets}
\end{table*}

\begin{table*}
\begin{tabular}{lrrrr}
\toprule
Category & Count &  Median &  0.25 &  0.75 \\
\midrule
Noun Phrase, Adjective-Animate &      2 &     0.0 &   0.00 &   0.00 \\
Determiner-Possessive &     12 &     2.0 &   2.00 &   2.00 \\
Verb-Transitive Phrase, Verb-Intransitive, Preposition Phrase &      2 &     2.0 &   2.00 &   2.00 \\
Determiner-Numeral Phrase &      2 &     2.0 &   2.00 &   2.00 \\
Determiner-Numeral &     36 &     2.0 &   2.00 &   3.75 \\
Altered POS, Determiner-Numeral &      2 &     2.0 &   2.00 &   2.00 \\
Verb-Intransitive Phrase, Adverb-Animate &      2 &     2.0 &   2.00 &   2.00 \\
Noun, Adjective-Size &      2 &     2.0 &   2.00 &   2.00 \\
Noun, Preposition Phrase, Scope &      2 &     2.0 &   2.00 &   2.00 \\
Adverb-Spatial &      2 &     2.0 &   2.00 &   2.00 \\
Adjective-Weight &      6 &     2.0 &   2.00 &   4.00 \\
Preposition &     42 &     2.0 &   2.00 &   4.00 \\
Adjective-Age &      4 &     2.0 &   2.00 &   3.50 \\
Adjective-Speed Phrase, Verb-Intransitive &      2 &     2.0 &   2.00 &   2.00 \\
Verb-Intransitive, Noun &      4 &     2.0 &   2.00 &   2.25 \\
Adjective-Size &     24 &     2.0 &   2.00 &   4.00 \\
Adjective-Shape &     12 &     2.0 &   2.00 &   2.50 \\
Scope, Adjective-Manner &      8 &     2.0 &   2.00 &   2.00 \\
Verb-Transitive &      8 &     2.0 &   2.00 &   4.00 \\
Verb-Transitive Phrase &      2 &     2.0 &   2.00 &   2.00 \\
Adjective-Color (3-way swap) &      2 &     2.0 &   2.00 &   2.00 \\
Adjective-Animate &     16 &     2.0 &   2.00 &   4.00 \\
Preposition Phrase &     12 &     2.0 &   2.00 &   4.00 \\
Verb-Intransitive, Determiner-Numeral &      2 &     2.0 &   2.00 &   2.00 \\
Noun Phrase, Adjective-Color &      2 &     2.5 &   2.25 &   2.75 \\
Noun, Verb-Intransitive &      2 &     2.5 &   2.25 &   2.75 \\
Pronoun, Verb-Intransitive &      2 &     2.5 &   2.25 &   2.75 \\
Scope, Altered POS, Verb-Intransitive, Verb-Transitive &      2 &     2.5 &   2.25 &   2.75 \\
Noun Phrase, Determiner-Possessive &      4 &     2.5 &   2.00 &   3.50 \\
Negation, Noun Phrase, Preposition Phrase &      2 &     2.5 &   2.25 &   2.75 \\
Noun &    214 &     3.0 &   2.00 &   4.00 \\
Scope, Noun, Preposition &      2 &     3.0 &   2.50 &   3.50 \\
Scope, Conjunction Phrase &      4 &     3.0 &   2.50 &   3.50 \\
Scope, Conjunction &      2 &     3.0 &   2.50 &   3.50 \\
Adjective-Color &     88 &     3.0 &   2.00 &   4.00 \\
Scope, Adjective-Texture &      2 &     3.0 &   3.00 &   3.00 \\
Verb-Intransitive, Verb-Transitive Phrase &      4 &     3.0 &   2.50 &   3.00 \\
Sentence &     10 &     3.0 &   2.00 &   3.50 \\
Adjective-Spatial &      4 &     3.0 &   2.50 &   3.50 \\
Noun Phrase &     44 &     3.0 &   2.00 &   4.00 \\
Pronoun, Noun Phrase &      2 &     3.0 &   3.00 &   3.00 \\
Adjective-Temperature &     12 &     3.0 &   2.00 &   4.00 \\
Adjective-Temporal &      2 &     3.0 &   3.00 &   3.00 \\
Adverb-Animate &      2 &     3.0 &   2.50 &   3.50 \\
Verb-Intransitive, Adjective-Manner &      2 &     3.0 &   2.50 &   3.50 \\
Verb-Intransitive Phrase, Preposition &      2 &     3.0 &   2.50 &   3.50 \\
Verb-Intransitive Phrase &      2 &     3.0 &   2.50 &   3.50 \\
Scope, Preposition &      4 &     3.0 &   2.50 &   3.50 \\
Negation, Scope &     22 &     3.0 &   2.00 &   4.00 \\
\bottomrule
\end{tabular}

\caption{\textbf{ Alignment Scores for WinoGround}. The alignment columns reports the median and quartiles. The WinoGround results are the subset of groups with at least 10 examples. Plot continues on the next page.} 
\label{app:tbl:other-datasets}
\end{table*}

\begin{table*}
\begin{tabular}{lrrrr}
\toprule
Category & Count &  Median &  0.25 &  0.75 \\
\midrule
\texttt{WinoGround} 
Preposition Phrase, Scope &     14 &     3.0 &   2.00 &   4.00 \\
Scope, Preposition Phrase, Adjective-Color &      2 &     3.0 &   2.50 &   3.50 \\
Scope, Relative Clause &      2 &     3.5 &   2.75 &   4.25 \\
Verb-Transitive, Noun &      2 &     3.5 &   3.25 &   3.75 \\
Scope &      6 &     3.5 &   2.25 &   4.00 \\
Adverb-Temporal &      4 &     3.5 &   2.75 &   4.00 \\
Adverb-Spatial Phrase &      2 &     3.5 &   3.25 &   3.75 \\
Adjective-Speed &      4 &     3.5 &   2.75 &   4.25 \\
Scope, Preposition, Verb-Intransitive &      2 &     4.0 &   3.50 &   4.50 \\
Relative Clause, Scope &      2 &     4.0 &   4.00 &   4.00 \\
Determiner-Numeral, Noun Phrase &     10 &     4.0 &   4.00 &   5.00 \\
Altered POS &     36 &     4.0 &   2.00 &   4.00 \\
Adjective-Texture &     14 &     4.0 &   2.75 &   4.25 \\
Adjective-Manner Phrase &      2 &     4.0 &   4.00 &   4.00 \\
Adjective-Manner &     18 &     4.0 &   3.00 &   4.00 \\
Scope, Preposition Phrase &      4 &     4.0 &   3.00 &   4.00 \\
Noun, Adjective-Color &      2 &     4.0 &   4.00 &   4.00 \\
Scope, Verb-Transitive &      2 &     4.5 &   4.25 &   4.75 \\
Noun Phrase, Determiner-Numeral &      2 &     5.0 &   5.00 &   5.00 \\
Adjective-Height &      2 &     5.0 &   5.00 &   5.00 \\
\bottomrule
\end{tabular}

\caption{Continued.} 
\label{app:tbl:other-datasets-cont}
\end{table*}
\end{document}